\documentclass[10pt, conference]{IEEEtran}  % , a4paper, letterpaper
\IEEEoverridecommandlockouts % needed for the thanks command 
\usepackage{graphicx}
\usepackage{amsmath}
\usepackage{amssymb}
\usepackage[caption=false]{subfig}
\usepackage[utf8]{inputenc}
\usepackage{pgfplots}
\DeclareUnicodeCharacter{2212}{−}
\usepgfplotslibrary{groupplots,dateplot}
\usepackage{tikz}
\usetikzlibrary{shapes.geometric}
\usetikzlibrary{arrows.meta,arrows}
\usetikzlibrary{patterns,shapes.arrows}
\pgfplotsset{compat=newest}
\usepackage{environ}
\usepackage{multirow}
\usepackage{booktabs}
\usepackage{colortbl}

\usepackage{xcolor}
\usepackage[normalem]{ulem}
\definecolor{correction}{RGB}{255, 0, 0}

\definecolor{black}{RGB}{0, 0, 0}
\definecolor{orange}{RGB}{230, 159, 0} % stem dataset
\definecolor{skyblue}{RGB}{86, 180, 233}
\definecolor{green}{RGB}{0, 158, 115} % combined dataset 
\definecolor{yellow}{RGB}{240, 228, 66}
\definecolor{blue}{RGB}{0, 114, 178} % leaf dataset
\definecolor{red}{RGB}{213, 94, 0}
\definecolor{purple}{RGB}{204, 121, 167}
\definecolor{lightgray}{RGB}{204,204,204} % light gray for legend 

\usepackage{hyperref}       % hyperlinks
\definecolor{dark-red}{rgb}{0.4,0.15,0.15}
\definecolor{dark-blue}{rgb}{0.15,0.15,0.8}
\definecolor{medium-blue}{rgb}{0,0,0.5}
\hypersetup{
    colorlinks, linkcolor={dark-red},
    citecolor={dark-blue}, urlcolor={medium-blue}
}

\title{\LARGE \bf
Automated Phytosensing: Ozone Exposure Classification \\Based on Plant Electrical Signals
}
\author{Till Aust,$^{1}$ Eduard Buss,$^{1}$ Felix Mohr,$^{2}$ and Heiko Hamann$^{1}$%
\thanks{$^{1}$Till Aust, Eduard Buss and Heiko Hamann are with the Department of Computer and Information Science, University of Konstanz, Konstanz, Germany. 
{\tt\small \{till.aust, eduard.buss, heiko.hamann\}@uni-konstanz.de}}%
\thanks{$^{2}$Felix Mohr is with the Faculty of Engineering, Universidad de La Sabana, Ch{í}a, Colombia. 
{\tt\small felix.mohr@unisabana.edu.co}}}%
\begin{document}
\maketitle              % typeset the header of the contribution
\thispagestyle{empty}
\pagestyle{empty}

%%%%%%%%%%%%%%%%%%%%%%%%%%%%%%%%%%%%%%%%%%%%%%%%%%%%%%%%%%%%%%%%%%%%%%%%%%%%%%%%
\begin{abstract}
In our project \textit{WatchPlant}, we propose to use a decentralized network of living plants as air-quality sensors by measuring their electrophysiology to infer the environmental state, also called phytosensing. 
We conducted in-lab experiments exposing ivy (\textit{Hedera helix}) plants to ozone, an important pollutant to monitor, and measured their electrophysiological response.
However, there is no well established automated way of detecting ozone exposure in plants. 
We propose a generic automatic toolchain to select a high-performance subset of features and highly accurate models for plant electrophysiology. 
Our approach derives plant- and stimulus-generic features from the electrophysiological signal using the \textit{tsfresh} library. 
Based on these features, we automatically select and optimize machine learning models using \textit{AutoML}. 
We use forward feature selection to increase model performance.  
We show that our approach successfully classifies plant ozone exposure with accuracies of up to 94.6\% on unseen data. 
We also show that our approach can be used for other plant species and stimuli. 
Our toolchain automates the development of monitoring algorithms for plants as pollutant monitors. 
Our results help implement significant advancements for phytosensing devices contributing to the development of cost-effective, high-density urban air monitoring systems in the future.
%Submitting here: \url{https://ieee-ssci.org/?ui=ci-in-engineering-cyber-physical-systems} (Systems/Technologies for CI sounds good )
% Subjects: 
% main: Biosensors
% subs: Monitoring, sensor networks, intelligent decision making
%DEADLINE: 01.10.2024
%MAX: 6 pages + 1 page references 
\end{abstract}

%%%%%%%%%%%%%%%%%%%%%%%%%%%%%%%%%%%%%%%%%%%%%%%%%%%%%%%%%%%%%%%%%%%%%%%%%%%%%%%%
\section{Introduction}
Since decades city populations are increasing at a fast pace, a phenomenon called urbanization.\footnote{\url{https://unhabitat.org/wcr/}}
One effect is increased traffic volume, which negatively impacts air quality. 
For example, in Europe just under two thirds of air pollution is accounted to traffic\footnote{\url{https://www.eea.europa.eu/highlights/emissions-from-road-traffic-and}} and up to a third of European citizens are exposed to concentrations of air pollutants that exceed the European Union's air quality standards, especially due to particulate matter and ozone.\footnote{\url{https://www.eea.europa.eu/soer/2015/europe/urban-systems}}
Hence, in addition to ongoing efforts to reduce pollution, also suitable air pollution monitoring techniques are needed to identify potential hazards early and to enable swift action~\cite{Ullo2020}. 
Usually, such techniques incorporate expensive, fixed monitoring stations that provide highly accurate measurements with the drawback of low spatial resolution~\cite{Xie2017}. 
Hence, there is a paradigm shift towards more decentralized measurement approaches but a fundamental challenge remains in finding cost-effective methods to monitor all pollutants accurately~\cite{Snyder2013}.
Another approach addressing low spatial resolutions is to use mobile pollution sensing systems. 
However, the increased spatial resolution comes with a tradeoff of reduced temporal resolution.

In our EU-funded project \textit{WatchPlant}~\cite{Garcia-Carmona2021,Hamann2021} (2021-2024) we address both the challenge of measuring many pollutants in an inexpensive and effective way and the tradeoff of temporal and spatial resolution by proposing a network of low-cost biohybrid air-quality sensor nodes. 
Our vision is to deploy these air-quality sensor nodes densely throughout the city (possible due to their low-cost) and connecting all of them to aggregate pollution information for real-time monitoring to help policy makers to decide and inform citizens about potential hazards, for example, via their smart phones. 
By engaging citizens and avoiding private data collection (e.g., connection history) we mitigate ethical concerns. 
We anticipate a positive ecological impact as our biohybrid system promotes the expansion of urban green spaces.
One biohybrid air-quality sensor node consists of an electronic device, our \textit{PhytoNode}~\cite{Buss2022,Buss2024}, and a natural plant (in our case \textit{Hedera helix}).
The \textit{PhytoNode} is a light-weight and inexpensive sensor used for phytosensing.
In phytosensing the electric physiology of living plants is measured to sense, for example, the environmental conditions~\cite{Pfotenhauer2024}. 
One advantage of phytosensing is that we can potentially infer multiple pollutants, which usually requires multiple different sensing devices, by only measuring one natural plant.
A~cost-effective and robust phytosensing approach is to measure electric differential potentials holistically across the entire plant. 
This can be done, for example, by inserting two needles as electrodes, one close to a leaf and one close to the bottom of the plant's stem. 
A~remaining challenge is to infer the environmental stimulus based on the measured holistic electric differential potential because this macroscopic level of electric physiology in plants is neither well studied nor thoroughly understood as of now~\cite{Sukhova2017}. 

Due to the recent successes of machine learning (ML) across many domains~\cite{Jordan2015,LeCun2015}, also several studies applied ML to analyze holistic plant electrophysiology. 
Most studies utilize phytosensing in the context of agriculture, for example, to detect drought stress using support vector machines~\cite{Tian2023}, ensemble boosted tree classifiers~\cite{Sai2023}, or (extreme) gradient boosting~\cite{Najdenovska2021,Tran2019}. 
Other studies utilize deep neural networks to detect salt stress~\cite{Qin2020,Yao2021}, nitrogen deficit~\cite{Gonzalez_i_jucla2023}, soil moisture~\cite{Qi2024} or other abiotic stressors~\cite{Pereira2018} (classical machine learning approaches for abiotic stressors~\cite{Sai2022,Buss2023}). 

% ozone response looks exactly like ours + exactly the other way around as dolfi
Fewer studies consider phytosensing in the context of air pollution monitoring, by experimenting with gas stimuli~\cite{Chaparro2021}. 
\textit{Chatterjee}~et~al.~\cite{Chatterjee2015} experimented with tomato plants and exposed them, among other stimuli, to ozone (16~ppm, that is, 266~times the threshold for preserving the citizen health stated by the EU\footnote{threshold value for ozone of 60~ppb (parts per billion), e.g. German Federal Environment Agency: \url{https://www.umweltbundesamt.de/daten/luft/ozon-belastung\#zielwerte-und-langfristige-ziele-fur-ozon}}) for one minute every 2~hours.
\textit{Chatterjee}~et~al. propose to use discriminant analysis classifiers to classify ozone pollution based on 11~features, with background subtraction, derived from the plant electrical signal. 
Depending on the selected features and classifiers, they can distinguish between ozone and other stimuli (different concentrations of sodium chloride and sulfuric acid) with an accuracy of up to 95~\%. 
This work has been extended by using more statistical features~\cite{Chatterjee2017}, curve fitting coefficients as features~\cite{Chatterjee2018}, or different classifiers~\cite{Bhadra2023}.
However, they did not study whether a stimulus can be distinguished from the resting state (i.e., no stimulus).

\textit{Dolfi}~et~al.~\cite{Dolfi2015} propose a two step detection algorithm to assess ozone air pollution based on plant electrical signals. 
In a first step, they use a derivative-based algorithm for change detection in the signal. 
If a change is detected, they employ correlation waveform analysis to assess if the change is due to ozone pollution. 
They exposed \textit{Ligustrum texanum} and \textit{Buxus macrophilla} to either ozone concentrations of 200~ppb or gradually increasing ozone levels from 50~to~200~ppb. 
They achieve an overall detection accuracy of 87~\%. 
However, their method requires to manually select three plant specific thresholds for the detection algorithm and plant specific waveform responses for the correlation measure. 
They also did not study if this methods generalizes to other pollutants or stimuli.
% TODO should we include: Soybean electrophysiology: effects of acid rain (http://www.esalq.usp.br/lepse/imgs/conteudo_thumb/Soybean-electrophysiology-effects-of-acid-rain.pdf)

All the above studies have in common that they choose the ML models ad hoc without providing a rationale for choosing one over another. 
They appear to have selected a functioning model through manual parameter adjustments or by comparing a restricted set of seemingly randomly chosen models. 
As a possible solution we propose to use automated machine learning~(AutoML). 
Methods of AutoML automatically compose and parameterize ML algorithms to optimize a provided metric~\cite{Yao2018}.
There are implemented frameworks, such as, auto-sklearn~\cite{Feurer2015, Feurer2020} and GAMA~\cite{Gijsbers2019}. 
More recently \textit{Mohr}~and~\textit{Wever}~\cite{Mohr2022} proposed the Naive AutoML framework which yields similar performance while significantly reducing computation time.

We aim to close the gap of automatically finding a model that can classify external stimuli based on (holistic) plant electric differential potentials to contribute to decipher the plant physiology. 
We propose a~toolchain based on:  (1)~\textit{tsfresh}~library~\cite{Christ2018} for generic feature extraction, (2)~Naive AutoML~\cite{Mohr2022} for automatically optimizing an ML pipeline that finds both the best model for a dataset and the best preprocessing steps, and (3)~adapted forward feature selection to find the most important features.
We apply this toolchain to \textit{Hedera helix} electrical differential potentials (recorded using our \textit{PhytoNode}), while the plant is exposed to ozone. 
We show that our toolchain also works on different plant species and other stimuli using data from \textit{Buss}~et~al.~\cite{Buss2023}.

In summary, we contribute to implementing low-cost air-pollution monitoring systems based on phytosensing by showing that 
\begin{itemize}
    \item we can classify ozone exposure in plant physiology in a new plant (\textit{Hedera helix}) with high accuracy;
    \item developed low-cost hardware (\textit{PhytoNode}) is capable of capturing the relevant signal dynamics necessary for classifying pollutants exposure; and
    \item our approach can generalize to other plant species and stimuli.
\end{itemize}
These advances in phytosensing technology, in combination with the present or future urban landscaping, would potentially allow for large-scale outdoor deployment of inexpensive and robust air-quality sensors to effectively monitor air pollution in urban areas. 
To ensure reproducibility, we have publicly released the code.\footnote{\url{https://github.com/tilly111/watchplant_classification}}

\section{Experiments and Methodology}
\label{sec:experiments_and_methods}
Next, we describe the experimental setup (Sec.~\ref{sec:experimental_setup}), the obtained data, and how we preprocessed it. 
We describe our toolchain, consisting of deriving plant species and stimulus generic features from the electric differential potential using the \textit{tsfresh} library (Sec.~\ref{sec:data_porcessing_and_feature_extraction}), followed by using these generic features to find and optimize a machine learning model using the Naive AutoML framework (Sec.~\ref{sec:automated_model_finding}), and finally employing an adapted forward feature selection to find good working feature subsets (Sec.~\ref{sec:feature_selection}). 

\subsection{Experimental setup}
\label{sec:experimental_setup}
\begin{figure}[t]
  \centering
  \input{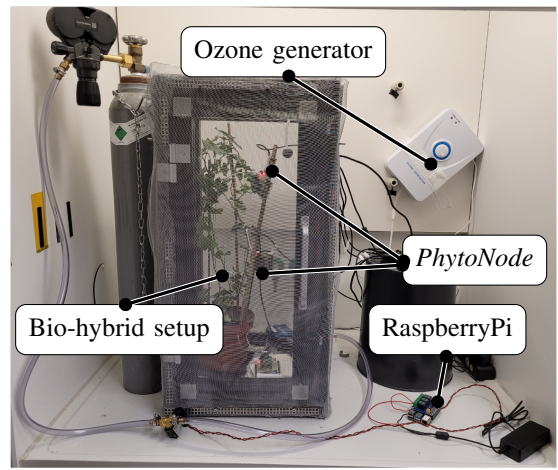}
  \caption{Experimental setup to expose an ivy plant to ozone. The ivy plant is within a Faraday cage to minimize external disturbances. We use two \textit{PhytoNode}s~\cite{Buss2024} to measure the plant electric differential potential and a RaspberryPi to store the data. Ozone is generated and induced from the top. The ozone concentration is controlled via two ozone sensors (one at the top and one at the bottom).}
  \label{fig:gas_setups}
\end{figure}
In Fig.~\ref{fig:gas_setups} we show the experimental setup for the ozone experiments~\cite{Buss2024}. 
We use two \textit{PhytoNode}s to simultaneously measure the electric differential potentials close to the leaves and roots with a frequency of approximately 300~Hz. 
Each \textit{PhytoNode} has a pair of electrodes that are inserted approximately 15~cm apart into the plant's tissue. 
To reduce external disturbances (e.g., electrical noise and static charges), we measure inside a Faraday cage. 
The ivy plant is exposed to 10~minutes of increased ozone concentration (peak exposure mean 1,447~ppb~$\pm$~376~ppb; no exposure mean {$-14$}~ppb~$\pm$~52~ppb, negative values due to calibration of the sensors). 
Then we allow the plant to recover for 2~hours before iterating and exposing it to ozone again. 
We exposed four different ivy plants to ozone for 17, 19, 18, and 24~repetitions respectively. 
This yields a total number of 78 \emph{expositions}. 

For the class of \textit{no ozone} (i.e., no increased ozone application) we select the 10~minutes right before the ozone exposition. 
Hence, we have a fully balanced dataset of 156~ten~minute slices of electric differential potential of ivy plants, 78 instances with and 78 instances without ozone exposition.

The measurements of the expositions are organized in three datasets.
One with the signals received in the plant leafs (\emph{Leaf} dataset), and one with the signals received in the plant stem (\emph{Stem} dataset).
We also consider the combined (and time-wise aligned) data (\emph{Combined} dataset).

From these datasets, we remove measurements of those expositions in which the majority of the signal was not recorded. 
This is the case for 2~expositions (4 samples) in both the leaf and stem measurements of the first plant and 15~expositions (30 samples) of the stem measurements of the second plant. 
So we have 152~samples of leaf measurements, 122~samples of stem and combined measurements.

We split each of the three datasets (\emph{Leaf}, \emph{Stem}, and \emph{Combined}) randomly into an analysis dataset consisting of 80\% of all the data (across plants) and a test dataset consisting of the remaining 20\%. 
For the subsequent processing and analysis, only the analysis datasets are used, if not stated explicitly otherwise.

\subsection{Data preprocessing and feature extraction}
\label{sec:data_porcessing_and_feature_extraction}
\begin{figure}[t]
  \centering
  \input{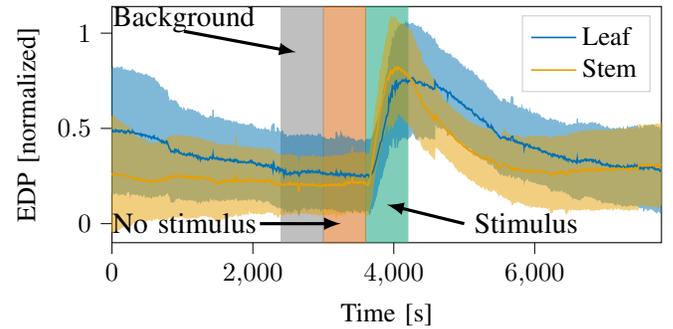}
  \hspace*{-0.2cm}
  \caption{Example data of the first experiment with 17~expositions. The solid lines show the mean scaled electric differential potential (EDP) over all expositions of the ivy plant (blue at the leaf, orange at the stem). The shaded areas give the standard deviation. The green time interval shows the ten minute slice selected for calculating the stimulus features, the red time interval shows the ten minute slice selected for calculating the no stimulus features, and the gray time interval indicates the ten minute slices used for background subtraction.}
  \label{fig:example_data}
\end{figure}

First, both leaf and stem measurements are converted from raw sensor readings to millivolt~(mV). 
This transformation allows to eliminate all physically illogical values by removing all electric differential potential measurements that exceed a threshold of $\pm$~200~mV. % this step is basically not required using our measurement device 
Next, we employ a preprocessing pipeline inspired by \textit{González I. Juclà}~et~al.~\cite{Gonzalez_i_jucla2023} that includes a rolling median filter with window size~10, sampling the data down to 2~Hz, and finally cutting the data into 10~minute slices. 
From each exposition we use the 10~minutes during which the stimulus was applied, the 10~minutes directly before the stimulus application, and 10~to 20~minutes before stimulus application for background subtraction, see Fig.~\ref{fig:example_data}. 
From each 10~minute slice we calculate 787~features using the \textit{tsfresh}~library~\cite{Christ2018}. 
We chose \textit{tsfresh} because it describes generic time series features, which are applicable for finding a generic and robust solution.
Background subtraction is used to mitigate inter-plant variance~\cite{Buss2023,Chatterjee2015} and is calculated by 
\begin{align*}
    \text{feature}_\text{sub} = \text{feature}_\text{ozone or no ozone} - \text{feature}_\text{background}.
\end{align*} 
Hence, each sample of the \emph{Leaf} and \emph{Stem} dataset consists of 787~features. 
The \emph{Combined} dataset consists of 1,502~features as we remove constant features (i.e., same value across all samples).

\subsection{Automated model finding}
\label{sec:automated_model_finding}
We use the Naive AutoML library~\cite{Mohr2022} to find the optimal machine learning pipeline for our data in a binary classification task using all features.
This library searches in \textit{sci-kit learn}~\cite{scikit-learn} classifier and preprocessing implementations for the classification pipeline that minimizes or maximizes a given metric.
Optimization is greedy in two phases by (1)~finding the best algorithm combination of pre-processors and classifiers under default hyperparameters through enumeration and (2)~optimizing the hyperparameters of the selected algorithms through random search (e.g., randomly selecting hyperparameters until a valid configuration is found in the default hyperparameter search space of Naive AutoML).
Once a pre-defined number of optimization steps has been executed, the best found pipeline is trained on the full dataset and returned.
While the greedy decisions could, in principle, lead to sub-optimal pipelines, it was shown empirically that the solutions generally have no or only a very small performance gap to solutions returned by time-intensive methods~\cite{Mohr2022}.

This library is easy to configure by allowing to set timeout limits, maximum number of hyperparameter iterations, or evaluation metrics.
We do not set a timeout limit but allow only 100~hyperparameter optimization steps (default parameter).  

For evaluating our model we run Naive AutoML once with each of two metrics.
One run was with the receiver operating characteristics area under the curve (ROC~AUC) as we have a binary classification problem, and it is important to incorporate uncertainty of the classifier into the evaluation. 
ROC~AUC captures this uncertainty.
In a second run, we optimized for accuracy (ACC), because it will ultimately be the essential measure for our envisioned application.
In both cases, Naive AutoML was configured to evaluate pipelines based on 5~independent and stratified 80\%/20\% training/validation splits.

\subsection{Feature selection}
\label{sec:feature_selection}
Based on the optimized classification pipeline, we investigate what subset of features performs best. 
Therefore, we do a semi-greedy approach, where we do 787 feature selection rounds (or 1,502 in the case of the \emph{Combined} dataset). 
Each round consists of the following steps: (1)~we sequentially iterate through all available features (not yet selected features), (2)~evaluate the performance for each choice (averaged over 100~independent runs), and (3)~add the feature to the set of selected features if it maximizes the performance. 
To allow for some temporary suboptimality, we keep a number of $n$~best selected feature sets for every feature set size. 
For each selected feature set, we iterate through all available features in the next round of feature selection. 
For the first 40~selected features we keep the 10~best feature sets; for selected features 41 to 100 we keep the 5~best feature sets; and after 100~selected features the 3~best feature sets. 
As performance measure we choose ROC~AUC, because it not only considers definite predictions but also scores the certainty of the classifier.

% On the most promising feature subsets we employ feature importance analysis using the \textit{SHapley Additive exPlanations}~(SHAP)~\cite{Lundberg2017} framework. 
% This framework provides a model agnostic, fair, and consistent explanation on how features contribute to specific predictions.
%The Shapely value determines how much each feature contributes to the difference between the actual prediction and the average prediction over all possible subsets of features.

\section{Results and Discussion}
\label{sec:results}
% short summary of what we do
%In this paper we present a toolchain that automatically generates generic features of plant electric differential potentials, using the \textit{tsfresh}~library. 
%These features are used to find and optimize a ML pipeline, using the Naive AutoML framework, for classifying the plant's exposure to ozone. 
%The optimized ML pipeline is then used to find the best subset of features and feature importance analysis is done.
%These findings help to foster a unified methodology for phytosensing especially in the field of air pollution monitoring which is essential for developing a biohybrid sensor network to monitor air-quality in urban areas.
Following the experimental protocol described in the previous section, we now describe the relevant intermediate and final results in three steps.
First, Sec.~\ref{sec:results:model_selection} details the results of the model selection phase from Naive AutoML.
Second, Sec.~\ref{sec:results:feature_selection} gives insights into the learning behavior of the selected pipelines in terms of feature and learning curves. 
Sec.~\ref{sec:results:generalizability} assesses selected models on test data and provides an unbiased estimator for the generalization performance.

\subsection{Model selection}
\label{sec:results:model_selection}
%We begin by generating 787~generic time series features per measured 10~minute signal, which we use to find the best classification pipeline for each analysis dataset (leaf, stem, and combined) using Naive AutoML. 
Table~\ref{tab:results_AutoML} summarizes the best found pipelines (preprocessing plus classifier) depending on the metric used for optimization (either ROC~AUC or accuracy).
The performance is reported as the average observed in the 5-fold hold-out validation as described in Sec.~\ref{sec:automated_model_finding}. 
\begin{table}[t]
    \centering
    \caption{Results of the automated model finding using Naive AutoML. RFC: Random Forrest classifier; kNN: k-nearest neighbor classifier; HGB: HistGradientBoosting classifier; GUS: Generic univariate select; ETC: Extra trees classifier.}
    \begin{tabular}{|c|c|c|c|c|}
    \hline
    \rowcolor{skyblue!50} \textcolor{black}{Dataset} & \textcolor{black}{Metric} &\textcolor{black}{Score} & \textcolor{black}{Best model} & \textcolor{black}{Preprocessing} \\
    \hline
    Leaf & Accuracy & 0.8880 & RFC & Variance Threshold \\
    Stem & Accuracy & 0.7800 & kNN & Min-max Scalar\\
    Combined & Accuracy & 0.9900 & HGB & None\\ % accuracy using all features 
    \hline
    Leaf & ROC AUC & 0.9654 & RFC & None \\
    Stem & ROC AUC & 0.8560 & ETC & Normalizer \\
    Combined & ROC AUC & 0.9333 & ETC & None \\
    \hline
    \end{tabular}
    \label{tab:results_AutoML}
\end{table}
We notice that ensemble classifiers (e.g., random forest classifier, extra tree classifier) work well, as they are often selected by Naive AutoML. See our implementation for the hyperparameter sets. 
However, the performance is generally similar among these models, and due to the small size of the training data, model selection is subject to significant noise and may change if optimization is repeated.
For larger datasets, we would expect more consistent model selection because noise and variability are mitigated. % For larger datasets we would expect more consistent model selection because noise and variability associated with small datasets are mitigated.

\begin{figure}[t]
    \centering
    \input{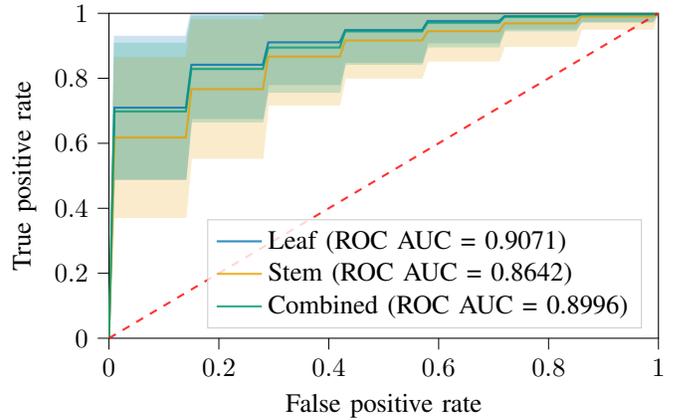}
    %\hspace*{-0.2cm}
    \caption{Each ROC curve is averaged (solid line) over 500 splits using 80\%/20\% stratified shuffle split and all features of the analysis datasets (blue: leaf, orange: stem, and green: combined). The shaded area corresponds to the standard deviation. The suggested pipeline from Table~\ref{tab:results_AutoML} is used for each setting.}
    \label{fig:roc_auc_curves}
\end{figure}

Next, Fig.~\ref{fig:roc_auc_curves} shows the ROC curves for the pipelines from Table~\ref{tab:results_AutoML}.
\begin{figure}[t]
  \centering
  \input{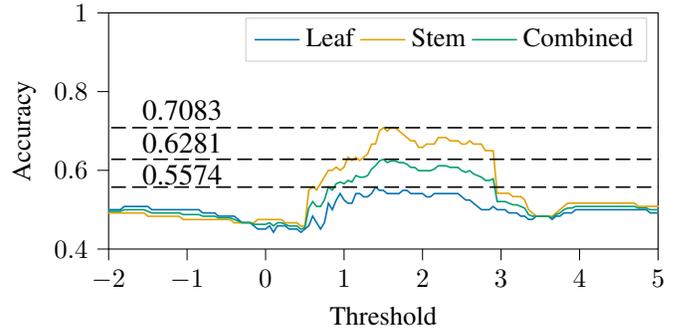}
  \caption{Accuracy results for a simple threshold model (blue: leaf, orange: stem, and green: combined). The model is evaluated using all experiment data and the result is averaged over 500~independent runs with a random 80\%/20\% train/validation split. See our implementation for details.}
  \label{fig:threshold_model}
\end{figure}
To ensure robust interpretation, the curves are obtained from 500 repeated evaluations of the pipeline with independent and stratified 80\%/20\% splits of the analysis data. 
One can see that classification is more successful using the leaf data (ROC~AUC~=~0.9071, ACC~=~82.15\%) rather than the stem signal (ROC~AUC~=~0.8642, ACC~=~77.51\%). 
Using both signals (e.g., information from the leaf and close to the root) yields a similar separability performance as solely using the leaf signal (ROC~AUC~=~0.8996, ACC~=~83.49\%).
We observe high variance in the results (shaded areas), which can be attributed to the small training (and validation) set sizes.

For comparison, we show a simple threshold model in Fig.~\ref{fig:threshold_model}.
%The signal is preprocessed, randomly split into 80\% training and 20\% validation data (results are averaged over 500~independent runs, see implementation for details) and we obtain a maximum accuracy of 70.83\% using the stem dataset. 
The signal is preprocessed, randomly split into 80\% training and 20\% validation data and we obtain a maximum accuracy of 70.83\% using the stem dataset. 
%Hence, it is not a trivial classification problem and the complexity can be explained by the fact that the electric differential potentials are on different levels depending on the daytime due to the plants system potential~\cite{Li2021,Zimmermann2016}. 
Hence, classification is not trivial and the complexity can be explained by varying electric differential potential levels depending on the daytime due to the plant's system potential~\cite{Li2021,Zimmermann2016}.
Possibly due to system complexity, plants may respond physiologically individually different to the same external stimulus.

\subsection{Features selection}
\label{sec:results:feature_selection}
Based on the pipeline that is best using \emph{all} features, we analyze which feature subsets to choose and if it can further improve the performance of the ML models.
This analysis is based on ROC~AUC.

\begin{figure}[t]
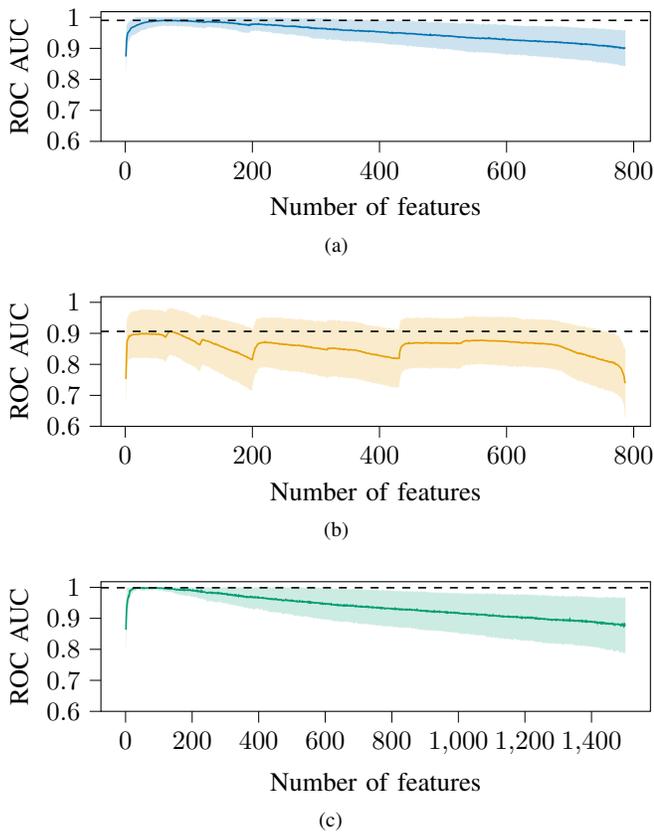

  \centering
  \subfloat[]{
    \input{images/final_figures/roc_auc_over_features_analysis_pn1}
    \label{fig:pn1_feature_subset_all}
    }\\
  \subfloat[]{
    \input{images/final_figures/roc_auc_over_features_analysis_pn3}
    \label{fig:pn3_feature_subset_all}
    }\\
  \subfloat[]{
    \input{images/final_figures/roc_auc_over_features_analysis_pn1_pn3}
    \label{fig:pn1_pn3_feature_subset_all}
    }
  \caption{Mean ROC AUC score (solid line) for the all feature feature subsets of the (a)~leaf, (b)~stem, and (c)~combined analysis dataset. 
  The best performance is reached using (a)~62~features (AUC ROC score of 0.9901) for the leaf analysis dataset, (b)~69~features (AUC ROC score of 0.9063) for the stem analysis dataset, and (c)~94~features (AUC ROC score of 0.9985) for the combined analysis dataset. Shaded area is the standard deviation.}
  \label{fig:feature_subset_all}
\end{figure}

In Fig.~\ref{fig:feature_subset_all} we show the ROC~AUC scores as a function of the feature set size, where feature sets are determined using the selection approach described in Sec.~\ref{sec:feature_selection}. 
The best score for the \emph{Leaf} (ROC~AUC~=~0.9901, ACC~=~94.60\%), \emph{Stem} (ROC~AUC~=~0.9063, ACC~=~82.64\%), and \emph{Combined} analysis dataset (ROC~AUC~=~0.9985, ACC~=~89.34\%) is reached using 62, 69, and 94~features, respectively. 
All results are averaged over 100~independent repetitions (analysis data were randomly split into 80\%/20\% training and validation data). 
Generally, good performance is reached with a small subset of features and gradually decreases for larger feature sets. 
For each dataset, using a smaller subset of features can increase the ROC~AUC score compared to using all features ($\Delta$ \emph{Leaf}: 0.0894, $\Delta$ \emph{Stem}: 0.0421, and $\Delta$ \emph{Combined}: 0.0989).  
This maybe counter-intuitive finding can be explained by the scarcity of training data, which is much smaller than the number of features generated with \textit{tsfresh}, which is also a possible explanation for the noncontinuous behavior of the stem feature subsets (Fig.~\ref{fig:pn3_feature_subset_all}).

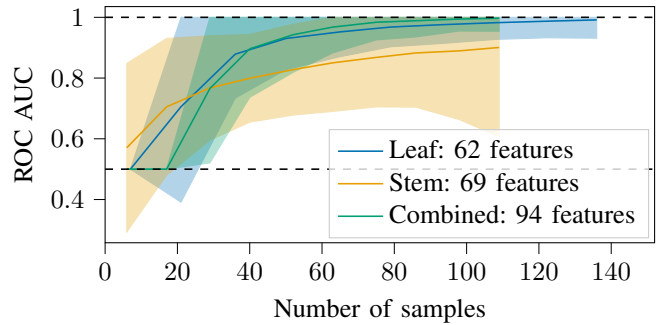
\begin{figure}[t]
  \centering
  % This file was created with tikzplotlib v0.10.1.
\begin{tikzpicture}

\begin{axis}[
width=.49\textwidth,
height=.2\textheight,
legend cell align={left},
legend style={
  fill opacity=0.8,
  draw opacity=1,
  text opacity=1,
  at={(0.99,0.25)},
  anchor=east,
  draw=lightgray
},
tick align=outside,
tick pos=left,
x grid style={darkgray176},
xlabel={Number of samples},
xmin=0, xmax=152,
xtick style={color=black},
y grid style={darkgray176},
ylabel={ROC AUC},
ymin=0.257673367075892, ymax=1.0353488872821,
ytick style={color=black}
]
\path [draw=blue, fill=blue, opacity=0.3]
(axis cs:7,0.5)
--(axis cs:7,0.5)
--(axis cs:21,0.391121930923911)
--(axis cs:36,0.73387292444574)
--(axis cs:50,0.827639077714472)
--(axis cs:65,0.870273579785061)
--(axis cs:79,0.902991988427503)
--(axis cs:93,0.913772541756761)
--(axis cs:108,0.927175348359074)
--(axis cs:122,0.932508740846806)
--(axis cs:136,0.931127036313408)
--(axis cs:136,1)
--(axis cs:136,1)
--(axis cs:122,1)
--(axis cs:108,1)
--(axis cs:93,1)
--(axis cs:79,1)
--(axis cs:65,1)
--(axis cs:50,1)
--(axis cs:36,1)
--(axis cs:21,1)
--(axis cs:7,0.5)
--cycle;

\path [draw=orange, fill=orange, opacity=0.3]
(axis cs:6,0.848098116012378)
--(axis cs:6,0.293022254357993)
--(axis cs:17,0.48004759247246)
--(axis cs:29,0.595695247652017)
--(axis cs:40,0.654674317198317)
--(axis cs:52,0.678046768808945)
--(axis cs:63,0.690137503217694)
--(axis cs:75,0.704975510022814)
--(axis cs:86,0.704085808000025)
--(axis cs:98,0.663119919934078)
--(axis cs:109,0.606544313717126)
--(axis cs:109,1)
--(axis cs:109,1)
--(axis cs:98,1)
--(axis cs:86,1)
--(axis cs:75,1)
--(axis cs:63,1)
--(axis cs:52,0.977360638598462)
--(axis cs:40,0.943671033845326)
--(axis cs:29,0.939341302055586)
--(axis cs:17,0.930438454039168)
--(axis cs:6,0.848098116012378)
--cycle;

\path [draw=green, fill=green, opacity=0.3]
(axis cs:6,0.5)
--(axis cs:6,0.5)
--(axis cs:17,0.5)
--(axis cs:29,0.520571001984864)
--(axis cs:40,0.736315043214526)
--(axis cs:52,0.818527080608642)
--(axis cs:63,0.882936956828578)
--(axis cs:75,0.925583690037798)
--(axis cs:86,0.935336265887886)
--(axis cs:98,0.954809781985926)
--(axis cs:109,0.954019707113968)
--(axis cs:109,1)
--(axis cs:109,1)
--(axis cs:98,1)
--(axis cs:86,1)
--(axis cs:75,1)
--(axis cs:63,1)
--(axis cs:52,1)
--(axis cs:40,1)
--(axis cs:29,1)
--(axis cs:17,0.5)
--(axis cs:6,0.5)
--cycle;

\addplot [semithick, blue]
table {%
7 0.5
21 0.70574546799852
36 0.878693518518518
50 0.930264957264957
65 0.951991830065359
79 0.96802380952381
93 0.975702898550725
108 0.982333333333333
122 0.987125
136 0.991476190476191
};
\addlegendentry{Leaf: 62 features}
\addplot [semithick, orange]
table {%
6 0.570560185185185
17 0.705243023255814
29 0.767518274853801
40 0.799172675521822
52 0.827703703703704
63 0.850172101449275
75 0.868011111111111
86 0.882657142857143
98 0.8896
109 0.90072
};
\addlegendentry{Stem: 69 features}
\addplot [semithick, green]
table {%
6 0.5
17 0.5
29 0.765895588235294
40 0.896299047619048
52 0.942715384615385
63 0.967825454545454
75 0.983772222222222
86 0.988759615384615
98 0.994333333333333
109 0.99704
};
\addlegendentry{Combined: 94 features}
\addplot [semithick, black, dashed, forget plot]
table {%
0 1
152 1
};
\addplot [semithick, black, dashed, forget plot]
table {%
0 0.5
152 0.5
};
\end{axis}

\end{tikzpicture} % ohne _2 einfache std
  \hspace*{-0.2cm}
  \caption{Learning curves for classifiers trained with (blue) 62~features of the leaf, (orange) 69~features of the stem, and (green) 94~features of the combined analysis dataset. Shaded area is three times the standard deviation. The results are averaged over 500~independent runs (80\%/20\% random training test split).}
  \label{fig:learning_curve}
\end{figure}

We plot the learning curves for all three datasets while using the best performing feature subset based on the analysis data in Fig.~\ref{fig:learning_curve}. 
All datasets show a logarithmic behavior, especially the \emph{Leaf} and \emph{Combined} dataset approximate 1.

The high variance (shaded areas are 3~standard deviations) in the \emph{Stem} learning curve can be likely attributed to the fact that the measurements are either generally less informative or more vulnerable to noise than measurements in the leaf.
While variance can generally also be attributed to a small dataset, the combined dataset has the same number of samples and does not show such a high variance.
Since the combined dataset has access to the leaf measurements, these are preferred over the stem measurements, which explains the more stable behavior of the combined curve.

The relatively high variance in all of the curves suggests that one must be prepared for significant deterioration of performance on the test data (and more generally on new data).
Specifically on the \emph{Stem} dataset, one can see that the variance increases towards the full dataset size, and since the final model will be trained on the full analysis data (beyond the curve end in the figure), even higher variance must be expected out of the analysis data.
Improvements in test data are theoretically possible but would be unexpected after a massive optimization process, which might have introduced meta overfitting on a small dataset.

% \subsection{Feature importance}
% \label{sec:results:feature_importance}
% % Leaf: 10 independent runs: ACC: 0.9538 pm 0.0510; ROC AUC: 0.9952 pm 0.0095
% % Stem: 10 independent runs: ACC: 0.8100 pm 0.1375; ROC AUC: 0.8960 pm 0.0916
% % Comb: 10 independent runs: ACC: 0.9300 pm 0.0781; ROC AUC: 1.0000 pm 0.0000
% % nicht so richtig sinnvoll weil wir generische feature genutzt haben 
% Based on the selected feature subsets, we do SHAP analysis, to see what features are most descriptive. 
% For the leaf dataset (using 62~features) \texttt{approximate\_entropy\_m\_2\_r\_0.3} was the most descriptive feature followed by \texttt{quantile\_q\_0.3}. 
% The two most descriptive features for the stem dataset (using 69~features) are \texttt{augmented\_dickey\_fuller\_attr\_pvalue\_autola g\_AIC\_} and \texttt{linear\_trend\_attr\_slope\_}. 
% Finally, the two most descriptive features for the combined dataset (using 94 features) are \texttt{fft\_coefficient\_attr\_abs\_coeff\_1} from the leaf dataset and \texttt{ar\_coefficient\_coeff\_1\_k\_10} from the stem dataset. 
% The nine out of ten most descriptive features of the combined analysis dataset are from the leaf dataset. 
% % Say something about the feature importance 
% We can constitute two findings, first, the most descriptive features do not align across the three datasets, and second, the SHAP values are all close to each other, i.e., the features are similar descriptive. 
% This makes sense, considering, that we used generic time series features.

\subsection{Generalizability}
\label{sec:results:generalizability}
\begin{figure}[t]
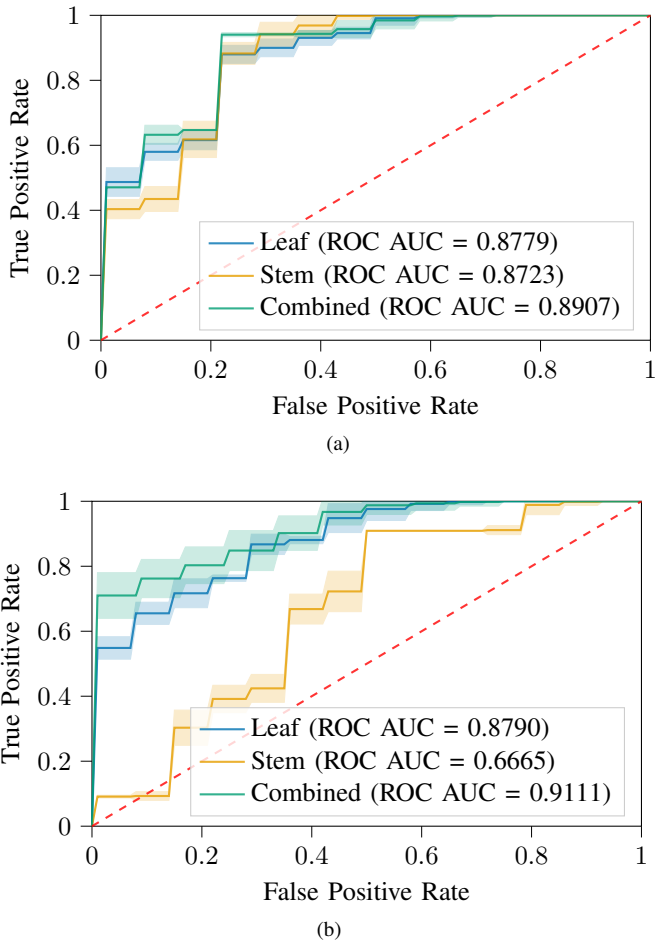

  \centering
    \subfloat[]{
        \input{images/final_figures/roc_curve_all_test.tex}
        \label{fig:roc_curve_test}
    }\\\hspace*{-0.2cm}
    \subfloat[]{
        \input{images/final_figures/roc_curve_all_test_feature_subset.tex}
        \label{fig:roc_curve_feature_subset}
    }
  \caption{ROC curves for the test data using (a)~all features (same setting as Fig.~\ref{fig:roc_auc_curves}) and (b)~the best subset found using the analysis data (62, 69, 94~features respectively, compare Fig.~\ref{fig:feature_subset_all}). Each curve is averaged over 500~independent runs where all analysis data was used for training and the test data are used for testing.} 
  \label{fig:feature_subset_all_test}
\end{figure}

Finally, we test how the found feature subsets and classifier pipelines perform on the test data. 
In Fig.~\ref{fig:roc_curve_test}, we show the ROC curves for all datasets using all features (similar setting to Fig.~\ref{fig:roc_auc_curves}). 
% analysis 
% Leaf: 0.9071 - 0.8779 = 0.0292
% Stem: 0.8642 - 0.8723 =-0.0081
% Comb: 0.8996 - 0.8907 = 0.0089
While the ROC~AUC slightly decreases for the \emph{Leaf} ($\Delta$: 0.0292) and \emph{Combined} ($\Delta$: 0.0089) dataset, it minimally increases for the \emph{Stem} dataset ($\Delta$: -0.0081). 
The accuracies decrease to 73.26\% (\emph{Leaf}), 74.19\% (\emph{Stem}), and 70.97\% (\emph{Combined}).  

In Fig.~\ref{fig:roc_curve_feature_subset} we show the ROC curves for all datasets using the suggested sub feature sets (see Fig.~\ref{fig:feature_subset_all}). 
% analysis
% Leaf: 0.8779 - 0.8790 =-0.0011  
% Stem: 0.8723 - 0.6665 = 0.2058
% Comb: 0.8907 - 0.9111 =-0.0204
While the average ROC~AUC score extremely decreases ($\Delta$:~0.2058) for the \emph{Stem} data, what could be expected due to the presented learning curve in Fig.~\ref{fig:learning_curve}, it slightly increases for the \emph{Leaf} ($\Delta$:~0.011) and \emph{Combined} ($\Delta$:~0.0204) data.  
The accuracies align and decrease to 57.21\% for the \emph{Stem} data, and slightly increase to 76.96\%, and 77.30\% for the \emph{Leaf} and \emph{Combined} data. 
The classification results with a high accuracy (up to 83.49\% or 94.60\% with feature selection on the analysis data) show that our proposed methodology is suitable for classifying ozone stress of ivy plants using their electrophysiology.
It also confirms that our developed \textit{PhytoNode} hardware is able to capture the relevant information in said signal.
The decline in accuracy on the test data can be explained by meta overfitting, that is, we overfitted the ML pipeline.
This also holds for the feature selection as it does not generalize reliable. 

The classification results show that measurements close to the leaf are better classifiable, thus exhibiting a clearer response, which aligns well with the intuition that ozone is absorbed by the leaves (as they `breathe' the ozone), resulting in a clearer and faster reaction in the plant.
However, it can be beneficial to measure plant electrophysiological signals at multiple locations. 
% What we cannot answer based on this dataset, because we only have two measurement positions, is where to ``optimally" place the electrodes for the case of multi-electrode measurements.

% showing generalizability 
\section{Application to other Species and Stimuli}
Finally and to assess the replicability of the proposed toolchain, we repeated the presented process on separate data of electric differential potential data of `ZZ' plants (\emph{Zamioculcas zamiifolia}, family Araceae) exposed to other stimuli, using different sensing technology (CYBRES phytosensor~\cite{CYBRES_UserManual})~\cite{Buss2023}.
While we spare details of the results for space reasons, we can report classifying accuracy between wind or no wind stimulus of 95.39\% using the extra trees classifier and no preprocessing step on unseen test data.
This indicates the generalizability of our toolchain. 

% % Not guranteed to find the best model
% Although, we use Naive AutoML, a tool that automatically optimizes a machine learning pipeline based on a given dataset, we cannot ensure to find the best working model nor is this approach deterministic, e.g., for the similar problem (same dataset) you can achieve different results. 
% However, we claim that this approach is more user friendly and finds better results faster than manually selecting and fine-tuning your models. 

\section{Conclusion and Outlook}
\label{sec:conclusion}
% What have we done 
We presented an automatic toolchain to classify external stimulus on a plant using the plant's electric differential potential signal. 
We propose to calculate generic time series features using the \textit{tsfresh}~library to be generic towards plant species and environmental stimuli. 
For finding a model, we propose to use Naive AutoML a tool that automatically finds and optimizes your machine learning pipeline.  
% What are the results 
The results show that this toolchain can be used to build and train models for different plant species and stimuli to achieve high accuracies. 
We conclude that, for ozone stress, the best response is measured close to the plant's leaves and that incorporating additional measurement points can enhance the detection capability. 
% The feature importance measures are inconclusive which is reasonable due to the selection of generic time series features.  

% What happens next
This generic toolchain allows for easily adjusting your prediction model to new plants as well as stimuli influencing the plant. 
With this in mind we can use this approach to prototype models which we can later use on the \textit{PhytoNode} to do real-time monitoring of air-pollution in cities using natural plants as sensors.

% What can we do next 
Next steps include integrating our findings into a real-world application, for example, onboard classification of ozone stress using the \textit{PhytoNode}. 
To ensure practical applicability, we must confirm that our models also work at lower ozone concentrations commonly encountered in everyday urban environments. 
We plan to include other relevant pollutant gases and start experimenting also on longer timescales (e.g., hours or days). 
The proposed automatic toolchain will simplify the integration of all these factors. 
We will consider also deep learning approaches; however, non-deep learning approaches seem more efficient and feasible in terms of space or computing complexity, considering a light-weight device, such as the \textit{PhytoNode}. 
We hope our toolchain will enable inexpensive scalable, real-time air-quality sensors and contribute to more effective and sustainable urban environmental monitoring.


%\section*{ACKNOWLEDGMENT}
\section*{Acknowledgment}

This work is supported by EU H2020 FET project WatchPlant, grant agreement No.~101017899 and the project ING-312-2023 at Universidad de La Sabana.
%%%%%%%%%%%%%%%%%%%%%%%%%%%%%%%%%%%%%%%%%%%%%%%%%%%%%%%%%%%%%%%%%%%%%%%%%%%%%%%%

\clearpage
\bibliographystyle{IEEEtran}
\bibliography{IEEEabrv,references}

% \section*{Appendix}
% \begin{figure}[h!]
%   \centering
%   \includegraphics[width=0.45\textwidth]{images/jaccard_index_stem.png}
%   \caption{Jaccard index for the stem dataset.}
%   \label{fig:jaccard_stem}
% \end{figure}

% \begin{figure}[h!]
%   \centering
%   \includegraphics[width=0.45\textwidth]{images/jaccard_index_stem_normalized.png}
%   \caption{Normalized Jaccard index for the stem dataset.}
%   \label{fig:jaccard_stem}
% \end{figure}

% \begin{figure}[h!]
%   \centering
%   \includegraphics[width=0.45\textwidth]{images/jaccard_index_leaf_normalized.png}
%   \caption{Normalized Jaccard index for the leaf dataset.}
%   \label{fig:jaccard_stem}
% \end{figure}

\end{document}